# High dimensional Bayesian Optimization Algorithm for Complex System in Time Series

Yuyang Chen, Kaiming Bi, Chih-Hang J. Wu, David Ben-Arieh, Ashesh Sinha

**Abstract**

At present, high dimensional global optimization problems with time-series models have received much attention from engineering fields. Bayesian optimization has quickly become a popular and promising approach for solving global optimization problems since it was proposed. However, the standard Bayesian optimization algorithm is insufficient to solving the global optimal solution when the model is high-dimensional. Hence, this paper presents a novel high dimensional Bayesian optimization algorithm by considering dimension reduction and different dimension fill-in strategies. Most existing literature about Bayesian optimization algorithms did not discuss the sampling strategies to optimize the acquisition function. This study proposed a new sampling method based on both the multi-armed bandit and random search methods while optimizing the acquisition function. Besides, based on the time-dependent or dimension-dependent characteristics of the model, the proposed algorithm can reduce the dimension evenly. Then, five different dimension fill-in strategies were discussed and compared in this study. Finally, to increase the final accuracy of the optimal solution, the proposed algorithm adds a local search based on a series of Adam-based steps at the final stage. Our computational experiments demonstrated that the proposed Bayesian optimization algorithm could achieve reasonable solutions with excellent performances for high dimensional global optimization problems with a time-series optimal control model.

## 1. Introduction

The global optimization problem of complex dynamic systems with high dimensions in time series has attracted recent attention. Conventional global optimization techniques are effective and suitable for solving the low-dimensional systems with the nature of time-independent or dimensions-independent [1, 2]. For instance, Ali and Kaelo introduced some modifications of the particle swarm algorithm to realize the faster convergence effectively and then tested the proposed algorithms on some problems with dimensions up to 20 [3]. Arora and Singh improved a basic butterfly algorithm to better explore the search space and applied the improved algorithm for the optimization problems with 30 dimensions [4]. An adaptive genetic algorithm consisting of new crossover and mutation operators is proposed to well escape from local minimums; the proposed algorithm is employed to solve different optimization problems with dimensions up to 40 [5]. These algorithms help them to reach a good performance for the lower-dimensional systems. However, they may suffer from relatively long running time for high dimensional complex dynamic systems or usually get stuck in a local optimum.

Bayesian optimization also is a promising and powerful global optimization approach to optimize complex dynamic systems with low dimensions [6]. However, when using the Bayesian optimization approach to solve global optimization problems with high-dimensional objective functions, evaluating the objective function iteratively could be extremely expensive and often impossible. Global optimization will also be challenging for general Bayesian optimization algorithms when dealing with time-series complex systems such as time-dependent optimal control systems; complex fluid flows systems, disease epidemic models, etc. Several Bayesian optimization algorithms in the existing literature were proposed for handling the high-dimensional global optimization problems. For example, Moriconi *et. al.* proposed a high dimension Bayesian optimization algorithm by learning a nonlinear feature mapping to reduce the inputs' dimension to easily optimize the acquisition function in low-dimensional space [7]. Zhang *et al.* introduced a sliced inverse regression method to Bayesian optimization to learn the intrinsic low-dimensional structure of the objective function in high-dimensional space [8]. Li *et al.* developed a new method for high dimensional Bayesian optimization using a dropout strategy to minimize dimensions and optimize a subset of variables [9]. Rana *et al.* proposed a high dimension Bayesian optimization algorithm to solve the acquisition function with a flat surface by gradually reducing the length-scale of the Gaussian process [10]. This paper reduces the variable dimensions by reducing the length-scale during the Gaussian regression part, which is different from other high-dimensional Bayesian optimization algorithms we reviewed in this paper. It can only reduce the dimension of Gaussian processes rather than optimizing the acquisition function in a low-dimensional space, which may lead to an inaccurate solution while it does calculational efficiency. The above high dimension Bayesian algorithms take a significant amount of time to reconstruct the decision variables from low-dimensional space back into high-dimensional space at each iteration and then calculate the corresponding objective function value in high-dimensional space. Thus, these existing algorithms do not efficiently realize the dimension reductions, the solution times were not significantly improved.

Moreover, the applications of many existing Bayesian optimization algorithms focus on high-dimensional systems with time-independent or dimension-independent properties [11]. Those algorithms can perform well on global optimization when the variables in their studied systems are time-independent or dimension-independent. They may lack efficiency on time-dependent or dimension-dependent systems. The time series system is an example of system variables with time-dependent properties, which means that variable values influence the future variable values in the previous time [12]. Solving these systems with the embedded time-dependent variables will significantly increase the computational complexity and challenge in high-dimensional systems.

The optimal control problem of a dynamic epidemic system is a typical high-dimensional case with time series in its problematic nature. During the epidemic, the health organizations or agencies may take a series of measures to determine the level of controls for local outbreaks, e.g., shelter-at-home, social distancing, person-protection-equipment, vaccination, quarantine, disinfection, or regional closures. All these mitigation measures could have significant associated financial costs, directly or indirectly. Suppose health organizations or agencies do not control the epidemic. In that case, it may also cause inevitable economic consequences, such as workforce losses due to outbreaks, increased community healthcare costs, local business downturns, and declined related travels. Thus, determining optimal control strategies of the ongoing epidemic has created a trend toward balancing the corresponding financial cost of control and the epidemic progression. An epidemic frequently lasts for a few hundred days or even a couple of years. The controls or interventions for these epidemic models are time-dependent since the state variable values and control value at the next time are affected by the state values and control values from one or more previous time. They also could be high-dimensional since such systems contain hundreds of thousands of time epochs; each time epochs can be considered a dimension of the models. As the number of time epochs increases, especially when the model is complex and nonlinear, or the objective functions are possible nonconvex, the challenges relate to global optimization of such epidemic control systems are rising. When there are many state variables in the model, it is necessary to calculate the values of state variables at each time epoch and sum up each epoch's cost to evaluate the overall cost of a control strategy. It could be too time-consuming to evaluate the overall cost, even for a single control strategy.

According to the above problems, this paper innovatively proposes a Bayesian optimization algorithm based on dimension reduction and dimension fill-in (DR-DF BO algorithm). This algorithm effectively resolves the shortcomings from the most existing high dimensional Bayesian optimization algorithms and obtains remarkable performance improvements in solving the global optimization solutions for the high-dimensional optimization problems with the possible nonconvex objective function, complex model, and time-dependent variables. Compared with the existing literature about high dimensional Bayesian optimization algorithms and standard Bayesian optimization algorithms, the main contributions of this paper are listed as follows:

(1) This paper presents an improved Bayesian optimization algorithm. This proposed algorithm combines both dimension reduction and dimension fill-in strategies. In this manner, the algorithm can effectively solve the global optimal solution for the high-dimensional complex models with time-dependent or dimensions-dependent variables.
(2) The proposed DR-DF BO algorithm proposes a variable dimension reduction strategy when the variables in the studied system are time-dependent. The proposed algorithm also doesn't require reconstructing the variable dimensions into original dimensional space at each acquisition optimization iteration, which significantly reduces the computational effort.
(3) This paper proposes a new sampling strategy to optimize the acquisition function by utilizing the multi-armed bandit concept and random search. This sampling strategy helps the proposed DR-DF BO algorithm learn the history sampling information and effectively sample the better points at each iteration.
(4) This paper introduces five strategies for the dimension fill-in for the proposed DR-DF BO algorithm, which may provide more options to meet different system requirements for further applications' use. Several fill-in strategies are tested and compared in this paper to increase the final solution accuracy and study which is better for the researched system.

The remainder of this paper is organized as follows. Section 2 defines two application problems formulated as the optimal control epidemic models, including one deterministic SEIR model and one stochastic SIS model. Section 3 presents each part of the proposed DR-DF BO algorithm in detail. Then the numerical simulation experiments are conducted to evaluate the proposed algorithm's performance in Section 4. Finally, Section 5 provides the conclusions and discusses our future work.

## 2. Application problem formulation

This section provides two application models formulations of the global optimization problem. The reason for presenting the application problems here is because they will be used to demonstrate the effectiveness and efficiency of the proposed algorithm in the simulation part. To better prove the global optimization performance of the proposed algorithm, we select the application problems with two different systems: deterministic and stochastic. Both models consider optimal control strategies with time-series variables. The proposed high dimension Bayesian optimization algorithm is expected to efficiently and accurately solve the global optimal control strategy that minimizes the objective function and subjects to the epidemic control model.

Usually, the control measures are not defined as variables in the general SEIR and SIS epidemic models [13, 14]. However, when the outbreak starts, the health organizations or agencies tend to determine the disease intervention level to control the spread of the epidemic, such as vaccination, quarantine, disinfection, or wearing masks. Those are all considered as control measures affecting the contact rate of infective individuals [15]. This paper defines the control variable in general SEIR and SIS epidemic models to balance the control measures on mitigating the disease spread and relieving the government financial burden. The control variable in the models indicates the level of the control measure.

### 2.1. Deterministic SEIR epidemic optimal control model

The deterministic SEIR epidemic optimal control model can be defined as following representation:

$$\text{Min} \quad V(u) = \int_{t_0}^{t_f} C_1 I(t) + C_2 f(u(t)) \tag{1}$$

$$\text{s. t.} \quad \frac{dS(t)}{dt} = \tau - \beta S(t)I(t) - \tau S(t) \tag{2}$$

$$\frac{dE(t)}{dt} = \beta S(t)I(t) - (\tau + \alpha)E(t) \tag{3}$$

$$\frac{dI(t)}{dt} = \alpha E(t) - (\tau + \gamma)I(t) - u(t)I(t) \tag{4}$$

$$\frac{dR(t)}{dt} = I(t) - \tau R(t) + u(t)I(t) \tag{5}$$

$$S(t) + E(t) + I(t) + R(t) = 1 \tag{6}$$

where $t_0$ is the start time, $t_f$ is the final time, assume the start time $t_0 = 1$ in this paper; $C_1$ and $C_2$ are the financial cost due to null control and the financial cost associated with the level of control in each time epoch, respectively; $f(u(t))$ represents the cost function of control variable regarding time; $u = \{u(1), \dots, u(t_f)\}$ is the set of control variables across $t_f$ dimensions, $u(t) \in [u_l, u_u]$, $u_l$ and $u_u$ are the lower and upper bound of control strategies, respectively; $S(t), E(t), I(t), R(t)$ are the system state variables, represent the susceptible, exposed, infectious and recovered population rate at time $t$, respectively; The parameter $\tau$ is the rate of natural birth, the rate of natural death is assumed to be equal to the natural birth rate; $\beta$ is the contact rate between susceptible individual and infectious individual; $\alpha$ is the transfer rate from exposed individuals to infectious individual; $\gamma$ is the rate for recovery.

### 2.2. Stochastic SIS epidemic optimal control model

Consider the effective contact rate of infectious individuals $\beta$ in the deterministic epidemic model (i.e., Eqns. (2) and (3)); this parameter is constant. Then each infectious individual makes $\beta dt$ the effective contacts with other susceptible individuals during the time interval $[t, t + dt]$. Now we assume that the effective contact rate $\beta$ changes to a stochastic parameter $\bar{\beta}$ caused by certain stochastic environmental factors such as seasonal variations, climate change, air humidity, etc. In a stochastic model, the effective contact each infectious individual in the time interval $[t, t + dt)$ will make is assumed as:

$$\bar{\beta}dt = \beta dt + \sigma dB(t) \tag{7}$$

where $B(t)$ is a standard Brownian motion, Eq. (7) means that the stochastic contact rate is normally distributed with mean $\beta dt$ and variance $\sigma^2 dt$ [14].

The stochastic SIS epidemic optimal control model with a time-varying contact rate is supposed as:

$$\text{Min} \quad V(u) = \int_{t_0}^{t_f} C_1 I(t) + C_2 f(u(t)) \tag{8}$$

$$\text{s. t.} \quad \frac{dS(t)}{dt} = \tau - \beta S(t)I(t) + \gamma I(t) - \tau S(t) + u(t)I(t) - \sigma S(t)I(t)dB(t)/dt \tag{9}$$

$$\frac{dI(t)}{dt} = \beta S(t)I(t) - (\tau + \gamma)I(t) - u(t)I(t) + \sigma S(t)I(t)dB(t)/dt \tag{10}$$

$$S(t) + I(t) = 1 \tag{11}$$

where the definition of parameters in the model (8) – (11) are the same in model (1) – (6).

To study the capability in solving the optimal control strategy for dynamic and high-dimensional models, this paper assumed that the cost functions $f(u(t))$ of the control variable in both SEIR and SIS models is a possible nonconvex function expressed as:

$$f(u(t)) = 0.3|\sin(10u(t))| + 2.1|\sin(u(t))| + u^2(t) \tag{12}$$

## 3. The proposed DR-DF BO algorithm

The traditional Bayesian optimization algorithm only considers the surrogate model, acquisition function, and random sampling for the acquisition function optimization, which is only sufficient to optimize low-dimensional models. The traditional Bayesian optimization algorithm cannot perform well for high-dimensional models since the high-dimensional acquisition function's optimization is still challenging. Especially for the time series models, the complexity of the global optimization straightly raises, which affects the accuracy of the global optimization of the Bayesian optimization algorithm. Therefore, this paper proposes a highly accurate high dimension Bayesian optimization algorithm based on dimension reduction and dimension fill-in.

The proposed DR-DF BO algorithm mainly includes six steps: variable dimension reduction, surrogate model, acquisition function, sampling strategies, local search with a series of Adam-based steps, variable dimension fill-in. Each step is introduced in detail in this section.

*3.1. Variable dimension reduction*

When the model is high dimensions, i.e., with many control variables, solving the optimal solution is usually computationally intractable. For example, in this paper, the control variables $u(t) \in [u_l, u_u]$, where $t = t_0, \dots, t_f$ with $t_f$ dimensions. If $t_f$ is larger than hundreds or even thousands, solving a global optimal control strategy in such a high dimensional space will become almost impossible. Dimension reduction is an effective way to transfer the data from high-dimensional space into low-dimensional space while retaining the important or meaningful properties of the original model.

Most conventional dimension reduction approaches can effectively deal with dimension reduction for high-dimensional variables in many fields. The variables in those approaches' applications have a common property. The remaining dimensions are independent after dimension reduction, and the removed dimensions usually contain less meaningful information of the original variables. However, the control variables in the models under study follow a time-series correlation in this paper. The value of the control strategy at each time epoch is considered as one dimension of the optimal control problem, it means that the control strategy at each time period is dependent on the strategies from several previous epochs. The control value $\mu(t)$ at time $t$ will affect the model's state variables at time $t$, and then it will affect the future control values. Thus, the conventional dimension reduction approaches may not be well-suited to handle the variables with time-dependent nature.

The control variables of SEIR and SIS models under study have $t_f$ dimensions, and the proposed DR-DF BO algorithm will evenly select $d < t_f$ $(d \in Z^+)$ dimensions to realize the dimension reduction. **Fig. 1**. provides some examples that the DR-DF BO algorithm determines the reduced dimensions in time.

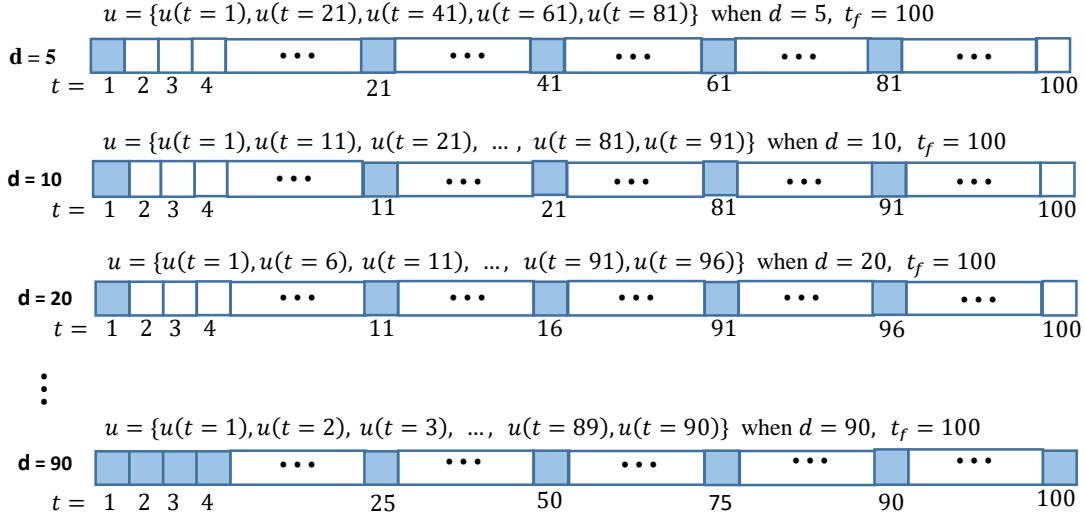

**Fig. 1.** The DR-DF BO algorithm selects the time dimensions for different $d$

*3.2. Surrogate model*

The surrogate model is an approximation model constructed by using a data-driven approach. There are many choices for many existing Bayesian optimization algorithms to build the surrogate model, for example, neural networks, random forests, and Gaussian Process. Because of the knowledge that the likelihood and the posterior is Gaussian Process if a function follows a Gaussian Process, the Gaussian Process develops as a popular choice of surrogate model for Bayesian optimization algorithm.

The Gaussian Process is a probability distribution over function. Herein, suppose the objective function $V(u)$ follows a Gaussian Process. Define $V(u^i)$ is the increment objective function value for $i$-th control strategy $u^i$, where $i \in Z^+$, $i$ is used to number the control, the control strategy with different $i$ only represent different control strategy. In here, simply denote the control strategy $u^i = \{u^i(1), \dots, u^i(\cdot)\}$ as $d$-dimensional variable after dimension reduction. Then there is:

$$V(u^i) \sim \mathcal{GP}(m(u^i), k(u^i, u^{i'})) \tag{13}$$

where $m(u^i)$ is called the mean function and $k(u^i, u^{i'})$ is the covariance function [16], where $u^i$ and $u^{i'}$ are two different control strategies. Oftentimes, the mean function is defined as either a linear function or directly defined as zero [17]. The covariance function is also named the kernel function.

Since the objective function follows the Gaussian Process, any finite number of the objective function values $V(u^i)$ follow multivariate Gaussian distribution [18]. Assume there are $p$ control strategies in total. Let $V = [V(u^0), \dots, V(u^i), \dots, V(u^p)]^T$ and $u^i$ is $i$-th control strategy, then $V$ is Gaussian distributed with mean vector $M = [m(u^0), \dots, m(u^p)]^T$ and covariance matrix $K$ as below:

$$K = \begin{bmatrix} k(u^0, u^0) & \cdots & k(u^1, u^p) \\ \vdots & \ddots & \vdots \\ k(u^p, u^0) & \cdots & k(u^p, u^p) \end{bmatrix} \tag{14}$$

For any new sampling point $u^j$ (a sampling point represents a $d$-dimensional control strategy) and the corresponding objective function value $V(u^j)$, let

$$V' = \begin{bmatrix} V \\ V(u^j) \end{bmatrix}, M' = \begin{bmatrix} M \\ m(u^j) \end{bmatrix}, \Sigma = \begin{bmatrix} K & K'^T \\ K' & K'' \end{bmatrix} \quad (15)$$

where $K' = [k(u^j, u^1), k(u^j, u^2), ..., k(u^j, u^p)]$, $K'' = k(u^j, u^j)$. Then the posterior distribution of $V(u^j)$ for any new sampling point $u^j$ based on the known dataset $V$ will Gaussian distributed with mean $\mu(V(u^j)|V)$ and variance $\sigma(V(u^j)|V)$, which can be written as:

$$V(u^j)|V \sim \mathcal{GP}(\mu(V(u^j)|V), \sigma(V(u^j)|V)) \quad (16)$$

where the posterior mean and the variance can be derived as:

$$\mu(V(u^j)|V) = m(u^j) + K'K^{-1}(V - M) \quad (17)$$

$$\sigma(V(u^j)|V) = K'' - K'K^{-1}K'^T \quad (18)$$

In the Gaussian Process, the kernel function, $k(u^i, u^{i'})$, is an important concept. The Bayesian optimization algorithm may perform different results with different kernel functions. Some popular kernel function choices include Matern32, Matern52, Radial Basis Function, Exponential, Linear, Brownian, Periodic, Polynomial, Warping, Coregionalize, RationalQuadrati [19]. For the proposed DR-DF BO algorithm in this paper, Matern52 is selected as the kernel function $k(u^i, u^{i'})$:

$$k(u^i, u^{i'}) = (1 + \sqrt{5} * \frac{|u^i - u^{i'}|}{l} + \frac{5}{3} * \frac{|u^i - u^{i'}|^2}{l^2}) \exp(-\sqrt{5} * \frac{|u^i - u^{i'}|}{l}) \quad (19)$$

where $l$ is the length-scale hyperparameter.

*3.3. Acquisition function*

The use of the acquisition function is an excellent way to define a trade-off between exploration and exploitation. The exploration will guide the algorithm to sample the next points that the objective function values could be highly uncertain. The exploitation will guide the algorithm to sample the next points leading to a lower objective function value. In this paper, the authors consider that the system is time-dependent, then the values of state variables at the current moment will affect their values at the next moment. When calculating the overall objective function value if the model contains many state variables, it is necessary to calculate the values of state variables at each time epoch and sum up each epoch's cost to evaluate the overall cost of a control strategy. It could be too time-consuming to evaluate the overall cost for a single control strategy. This is why the authors attempt to use the acquisition function to approximate the objective function for reducing the calculation effort and computational time.

Through the Gaussian Process, the acquisition function obtains the posterior mean and variance at a new sampling point $u^j$. The acquisition function can leverage the posterior distribution information to calculate a value that represents how desirable it is to sample next at this new point $u^j$. Some well-known choices of acquisition functions can provide a trade-off between exploration and exploitation for various applications. For example, lower confidence bound (LCB) [20], expected improvement (EI) [21], probability of improvement (PI) [22], Thompson sampling [23]. The original name of LCB is named Gaussian Process Upper Confidence Bound [24], it's originally proposed for the maximization problem. Since the optimization problem in this paper is a minimization problem, we use LCB here. PI is an alternative expression of EI. But PI is biased towards the exploitation over the exploration. When the variable is a single dimension, Thompson sampling usually performs better than other choices in running time.

This paper chooses LCB as the acquisition function of the proposed DR-DF BO algorithm. The expression is defined as:

$$\text{LCB}(u^j; k) = \mu(V(u^j)|V) - k\sigma(V(u^j)|V) \quad (20)$$

where $k$ is the weight to balance the posterior mean and variance. The purpose of the optimization problem is changed to find a control strategy, $u^j$, using the proposed DR-DF BO algorithm to minimize the Eq. (20).

*3.4. Sampling strategies*

Most of the current research works of Bayesian optimization do not discuss the sampling strategies for optimizing the acquisition function. The construction of the acquisition function is essential, but the sampling strategy to optimize the acquisition function also is the key to solve the global optimization problems efficiently. It does not efficiently handle the original optimization problem if we search the whole feasible solution space when optimizing the acquisition function. Random search is a simple and popular strategy to select the next sampling point for optimizing the acquisition function. However, if unlucky, the random search may either catch many similar sampling points that provide redundant information or never being able to locate the points closer to the global optima. This paper introduces the multi-armed bandit concept and considers the sampling problem as a multi-armed bandit problem.

The multi-armed bandit is a classic example of the exploration-exploitation trade-off. In the multi-armed bandit, at each time, the player decides to choose one or some machines from all machines, whether to continue with the current machine or try a different machine [25]. Each machine is configured with a reward probability of how you will likely earn a reward at each decision. In the proposed algorithm, we refer to different sampling value zones as different machines and refer to the number of sampling points at each value zone as the reward of the corresponding zone. The referred details will be presented in Section 3.4.1. The aim is to choose the best selection strategy at each time to achieve maximum long-term rewards. At each iteration, the decision logic is set up so that the algorithm can continuously gather more information to make better decisions later, which is referred to as the reinforcement learning process. Therefore, the DR-DF BO algorithm is proposed to combine the advantage of the multi-armed bandit and random search for the sampling process. It can deal with the optimization of the acquisition function effectively.

*3.4.1. Sample Points Based on Multi-Armed Bandit*

After dimension reduction, each control strategy should be $d$ dimensions. Therefore, the sampling point (a sampling point represents a control strategy) also is $d$ dimensions. Define the $d$-dimensional sampling point as $u^i = \{u^i(1), \ldots, u^i(\cdot)\} = \{u_1^i, \ldots, u_d^i\}$. The DR-DF BO algorithm is based on the multi-armed bandit that evenly divides the feasible region $[u_l, u_u]$ into $n$ separate zones. For each zone, the DR-DF BO algorithm samples $m$ points and there will be $m$ control strategies in each zone. It means for each iteration, each zone samples $m$ control strategies, each control strategy is $d$ dimensions, then there should sample $n*m$ control strategies in all $n$ zones. Therefore, at each zone, $m$ sampling points $u^1, \ldots, u^m \in \mathbb{R}^d$ were selected.

We consider the number of sampling points at each zone as the reward of this zone. So assume the initial reward of each zone is equal to $m$, then the $n$-dimensional reward matrix regarding all zones is $R = (m, m, \ldots, m)$. This reward matrix $R$ also represents the matrix of the number of sampling points in all zones. Calculate all acquisition function values corresponding to $n*m$ sampling points, find the sampling points with the smallest and largest acquisition function values and locate the corresponding zones these two sampling points belong to. We decide that the zone where the sampling point with the smallest acquisition function value belongs to receives a reward (for minimization problem), the zone where the sampling point with the largest value belongs to loses a reward. It means that in the next sampling iteration, one zone will be sampled with one less point, and one zone will be sampled with one extra point. This reward and reinforcement process can be generalized with other possible schemes. For example, we can define the zone where the sampling point with the second smallest value belongs to will earn $p$ rewards, the section where the sampling point with the largest value belong to will lose $q$ rewards, etc. Besides, among all $n*m$ sampling points, select the sampling point with the smallest acquisition function value as $u_M^*$, and express the corresponding acquisition function value as $y_M^*$.

*3.4.2. Sample Points Based on Random Search*

On the other hand, the DR-DF BO algorithm also takes advantage of random search to help with sampling points. The DR-DF BO algorithm specifies a uniform distribution with initial lower limit $L$ and initial upper limit $P$ for each dimension of control strategy $u$, the initial lower limit $L$ is equal to $u_l$, initial upper limit $P$ is equal to $u_u$. Then randomly sample value from the uniform distribution for each dimension of control strategy. In such case, the algorithm tries to use this sampling way to generate $N$ sampling points $u_R^1, u_R^2, \ldots, u_R^N$. Each sampling point is $d$ dimensions. Calculate all acquisition function values corresponding to these $N$ sampling points. Also, among all $N$ sampling points, the algorithm selects the sampling point with the smallest acquisition function value as $u_R^*$ and express the corresponding acquisition function value as $y_R^*$.

During each iteration of the DR-DF BO algorithm, the best $d$-dimensional control strategy can be selected as:

$$u^* = \begin{cases} u_M^*, & y_M^* < y_R^* \\ u_R^*, & otherwise \end{cases} \tag{21}$$

To more effectively sample points via the random search in each iteration, the DR-DF BO algorithm tries to update the lower limit and upper limit for the uniform distribution as:

$$L = \begin{cases} L + \alpha, & y_M^* < y_R^* \\ unchange, & otherwise \end{cases} \tag{22}$$

$$P = \begin{cases} P - \beta, & y_M^* < y_R^* \\ unchange, & otherwise \end{cases} \tag{23}$$

where $\alpha$ and $\beta$ are constant parameters. The values of $\alpha$ and $\beta$ in each iteration must satisfy the following condition:

$$u_l \leq L + \alpha < P - \beta \leq u_u \tag{24}$$

The purpose of updating the bound limit is to gradually and effectively narrow the feasible space, which utilizes the information gathered from the sampling process based on a multi-armed bandit. To better determine the values of $\alpha$ and $\beta$, we need to understand that $\alpha$ can be set as 0 or a smaller value if $u_M^*$ is closer to $u_l$, and $\beta$ can be set as 0 or a smaller value if $u_M^*$ is closer to $u_u$.

*3.5. Variable dimension fill-in*

Many existing high dimensional Bayesian optimization algorithms [7, 8, 10] consider reducing the variable's dimensions before optimizing the acquisition function. These algorithms find a low-dimensional sampling point with the optimal acquisition function value. Then the algorithms tend to transfer these low-dimension sampling points back to their original high-dimensional space to reconstruct a corresponding high-dimensional point. After that, they calculate the original objective function values corresponding to this high-dimensional solution space. Finally, the algorithms add this high-dimensional point and its objective function value into the database to update the surrogate model.

The existing Bayesian optimization algorithms do transform their models into a lower dimension space, however they transferred the decision variables back into their original dimension at each iteration. It results in not much computational time saving due to recalculating the original objective function values in its high-dimensional space. It is seen that these high dimensional Bayesian optimization algorithms do not successfully and efficiently realize the purpose of dimension reduction. Besides, it is unreasonable that those algorithms update the surrogate model using the high-dimensional reconstruction point and its corresponding objective function value. The surrogate model learns the posterior information and the relationship between the high-dimensional variables and objective function values. But the algorithms use that information to optimize the acquisition function value and find the low-dimensional point, which is inconsistent and will lower the solution's accuracy. That is why the DR-DF BO algorithm decides to do the variable dimension fill-in for the low-dimensional sampling point after the acquisition function's optimization process.

Considering the control strategy after the Adam-based local search process; we need to fill in the low-dimensional control strategy from the left $t_f - d$ dimensions to evaluate the objective function values in the whole space. The following section will introduce five different strategies to realize the variable dimension fill-ins.

**A. Identical value fill-in**

Identical value fill-in strategy means to fill in the left dimensions with the same values as the $d$ dimensions we obtain. We know that the control strategy after dimension reduction can be exactly expressed in Eq. (13), then the control values between time interval $[1, \lfloor \frac{t_f}{d} \rfloor + 1)$ can be filled with the same value $u(t = 1)$. To simplify, assume $\varphi = \lfloor \frac{t_f}{d} \rfloor$. The control values between time interval $[\varphi + 1, 2\varphi + 1)$ can be filled with the same value $u(t = \varphi + 1)$, and so on. **Fig. 2**. Shows the identical value fill-in process example when $d = 20$ and $t_f = 100$.

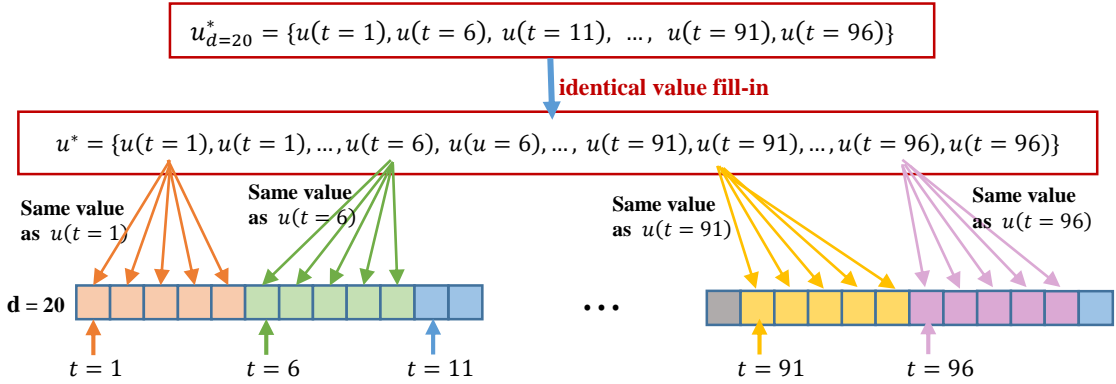

**Fig. 2.** Identical value fill-in strategy

**B. Uniform distribution fill-in**

Uniform distribution fill-in means to fill in the left dimensions by using the uniform distribution. For the $d$-dimensional control strategy:

$$u_d^* = \{u(t=1), u(t=\varphi+1), \ldots, u(t=(d-1)\varphi+1)\} \qquad (25)$$

The control values between time interval $\Delta = [q\varphi + 1, (q+1)\varphi + 1)$ can be filled by using the uniform distribution with the lower bound and upper bound as the following expression, respectively:

$$lower\ bound = \min(u(t=A), u(t=B)) \qquad (26)$$

$$upper\ bound = \max(u(t=A), u(t=B)) \qquad (27)$$

where $A = q\varphi + 1, B = (q+1)\varphi + 1$.

**C. Linear approximation fill-in**

Linear approximation fill-in means to fill in the left dimensions by using linear approximation approach. For the control values between time interval $\Delta$, we can approximate the control value using following equation and $2 \leq m \leq \varphi$:

$$u(t=A) + (m-1)\frac{u(t=B)-u(t=A)}{\varphi} \qquad (28)$$

**D. Normal distribution fill-in**

The normal distribution fill-in means to fill in the left dimensions by using a normal distribution. For the control values between time interval $\Delta$ can be filled by using the normal distribution with the mean and standard deviation of control strategies $u(t=A)$ and $u(t=B)$ as the following, respectively:

$$mean = mean(u(t=A), u(t=B)) \qquad (29)$$

$$std = std(u(t=A), u(t=B)) \qquad (30)$$

**E. Gaussian regression fill-in**

Gaussian regression fill-in means to fill in the left dimensions by using the Gaussian regression model [26]. For the $d$-dimensional control strategy as expressed in Eq. (25), the DR-DF BO algorithm learns the Gaussian regression model based on $d$ control values of this $d$-dimensional control strategy. For the control values between time interval $\Delta$, the algorithm uses the learned Gaussian regression model to predict the corresponding control values.

*3.6. Local search with Adam-based Steps*

The existing Bayesian optimization algorithms usually conclude that the algorithms find the final global optimal solution when they are done optimizing the acquisition function, we considered a local search with Adam-based steps in Bayesian optimization and reached a better final optimal solution in details in another paper [27]. To increase the solution's accuracy, the DR-DF BO algorithm decides to add a local search process based on a series of Adam steps after the acquisition function optimization, which can significantly improve the final solution quality in our computational experiments. Combining the steps of the DR-DF BO algorithm introduced above sections, the flow chart of the DR-DF BO algorithm is shown in **Fig. 3.** The complete implementation steps of the DR-DF BO algorithm are summarized in **Algorithm 1**.

**Algorithm 1** The DR-DF BO Algorithm
1: Initialize control strategy and state values of the model
2: Evenly select $d$ dimensions of the control strategy
3: Construct the Gaussian process model
4: **for** loop = 1, 2, … **do**
5:     **for** i = 1, 2, … $n$ sections **do**
6:         Evenly generate $m$ $d$-dimensional sampling points
7:         Find the best sampling point $u_M^*$ and corresponding acquisition function value $y_M^*$
8:     **end for**
9:     **for** j = 1, 2, … $N$ **do**
10:        Randomly generate $N$ sampling points $u_R^1, u_R^2, …, u_R^N$
11:        Find the best sampling point $u_R^*$ and corresponding acquisition function value $y_R^*$
12:     **end for**
13:     Find the best $d$-dimensional control strategy $u^*$ by comparing $y_M^*$ and $y_R^*$
14:     Check if need to update the lower limit $L$ and upper limit $P$
15:     Calculate $y^* = f(u^*)$, where $f$ is the original objective function
16:     Add the data $(u^*, y^*)$ into a database to update the Gaussian process model
17: **end for**
18: **obtain** the control strategy $u^*$ with best $f(u^*)$ recorded during iterations
19: **do** local search based on series of Adam-based steps starting from the point $u^*$
20: Fill in the point $u^*$ by using one of five strategies for dimension fill-in
21: **return** the final optimal control strategy and corresponding objective function value

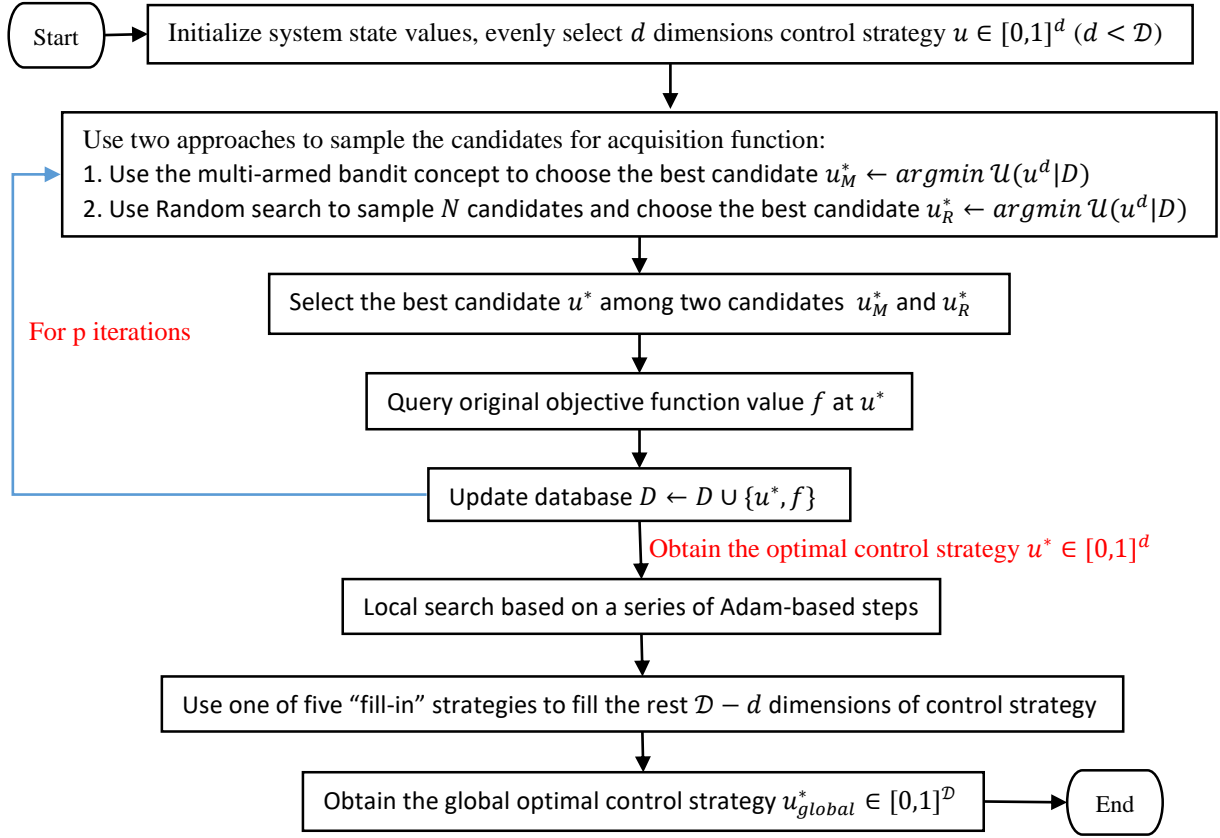

**Fig. 3.** The flowchart of DR-DF BO algorithm

## 4. Numerical Simulation

This section evaluates the proposed DR-DF BO algorithm's performance on two high-dimensional applications in time series and the comparisons with other algorithms. Moreover, the standard Bayesian optimization algorithm [28] and a high dimensional Bayesian optimization algorithm [9], are compared with the DR-DF BO algorithm proposed in this paper to show the advantages of the DR-DF BO algorithm. In this study, all simulation experiments are conducted on Python version 3.7 with Intel Core i5 CPUs and 32G memory, the Python libraries what we used includes torch, pyro, scikit-learn. The kernel function selected in the following simulation experiments is Matern52, and the lower confidence bound function is defined as the acquisition function.

*4.1. Global optimization on deterministic SEIR epidemic optimal control model*

This part verifies the global optimization performance of the DR-DF BO algorithm on the deterministic high-dimensional SEIR control model defined in Eq. (1) – Eq. (6), the global optimal control strategy is expected to not only control the epidemic spread but also minimize the overall financial cost. Under investigation, the deterministic SEIR model has the control variable with 100 time-epochs, so this problem's dimensionality is 100.

**Fig. 4** shows the SEIR epidemic model's comparison results with the control of a different dimension reduction and without any control when the fill-in strategy is a linear approximation. The solid red line represents the trend of infectious population rate without control over time. The other color lines represent the trend of the infectious population rate with optimal control generated by the DR-DF BO algorithm when the algorithm selects different low dimension values. As can be seen from **Fig. 4**, the infectious population increases sharply initially when the model is without control and declines very slowly to zero. However, no matter the low dimension's value, the optimal control strategies generated by the DR-DF BO algorithm effectively and quickly control the epidemic once it breakouts.

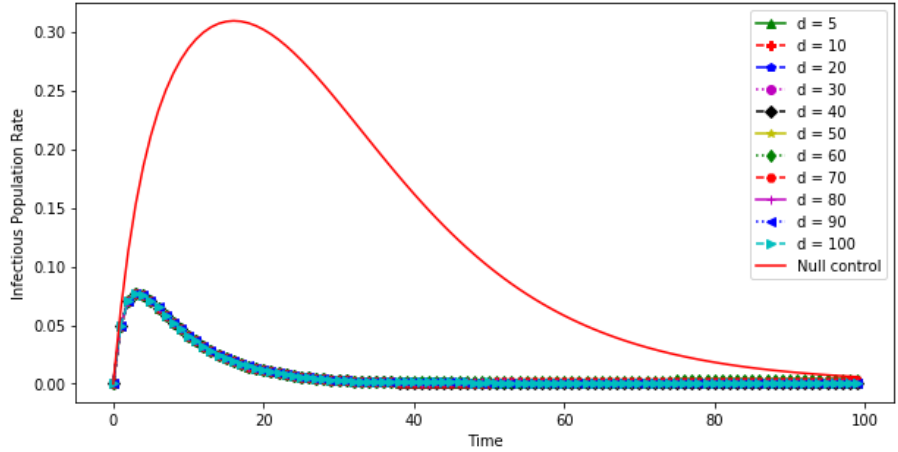

**Fig. 4.** Infectious population rate over time in deterministic SEIR model for different low dimension values

**Fig. 5** shows the trends of accumulated objective function value of the deterministic SEIR model with different low dimension values and without any control when the fill-in strategy is a linear approximation. The small figure in **Fig. 5** is a partial zoom figure between time interval [40,100]. We can see that the accumulated objective function when the model is without control is significantly higher than the values with control. Also, when $d$ value is about 40, the DR-DF BO algorithm can achieve a closer effect without dimension reduction (when $d = 100$). **Fig. 6** shows the final best objective function values and running time comparison results for different low dimension values when the fill-in strategy is a linear approximation. To make the results more intuitive, we use the ratio to express the results. We name the ratio of accumulated objective function value as AOFV ratio, and the ratio of running time as RT ratio, their expressions are defined as:

$$AOFV\ ratio\ (d) = \frac{AOFV(d)}{AOFV(d=100)} \tag{35}$$

$$RT\ ratio\ (d) = \frac{RT(d)}{RT(d=100)} \tag{36}$$

where $AOFV(d)$ and $RT(d)$ mean the accumulated objective function value and running time when dimension value is $d$, respectively.

The ratio results for the SEIR model are summarized in Table 1. The smaller the value of the ratio, the better the result. According to the simulation results, we can see that the DR-DF BO algorithm performs relatively well on both the final objective function value and running time when the dimension is reduced to 40. The proposed DR-DF BO algorithm can reach an excellent final solution using a reduced dimension of 40 with around 10 seconds of running time. The DR-DF BO algorithm shows an excellent global optimization performance for the deterministic SEIR model. It efficiently solves the optimal control strategy for the model to control the epidemic spread and significantly reduce the financial cost.

Table 1. AOFV ratio and RT ratio for different $d$ in SEIR model

| $d$ | 5 | 10 | 20 | 30 | 40 | 50 | 60 | 70 | 80 | 90 | 100 |
|---|---|---|---|---|---|---|---|---|---|---|---|
| $AOFV\ ratio$ | 1.218 | 1.218 | 1.218 | 1.1336 | 1.061 | 1.078 | 1.046 | 1.028 | 1.0 | 1.0 | 1.0 |
| $RT\ ratio$ | 0.222 | 0.251 | 0.333 | 0.425 | 0.517 | 0.652 | 0.667 | 0.754 | 0.821 | 0.908 | 1.0 |

**Fig. 7** shows the deterministic SEIR model's best objective values of different low dimension values when the DR-DF BO algorithm chooses different fill-in strategies. We can directly see that these five different fill-in strategies all provide good approximations on the tested global optimization problems. All of them can reach similar objective function values at $d = 40$, which indicates that the DR-DF BO algorithm can solve the optimal control solution within a reasonable running time no matter which fill-in strategy it uses.

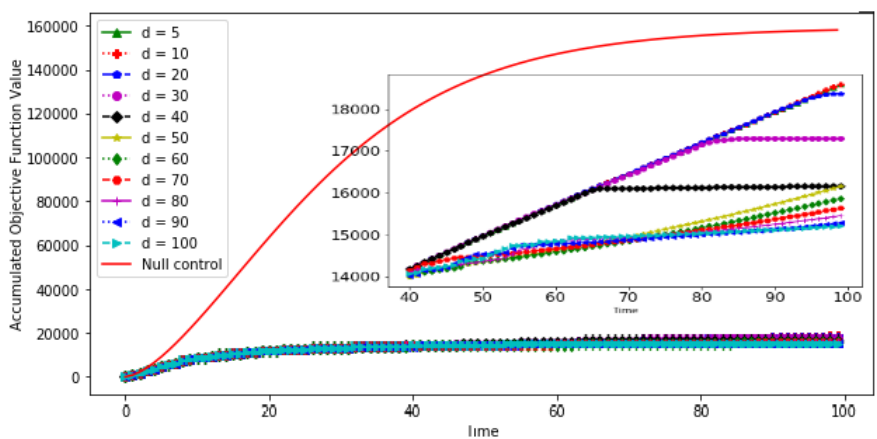

**Fig. 5.** Accumulated objective function values over time in deterministic SEIR model

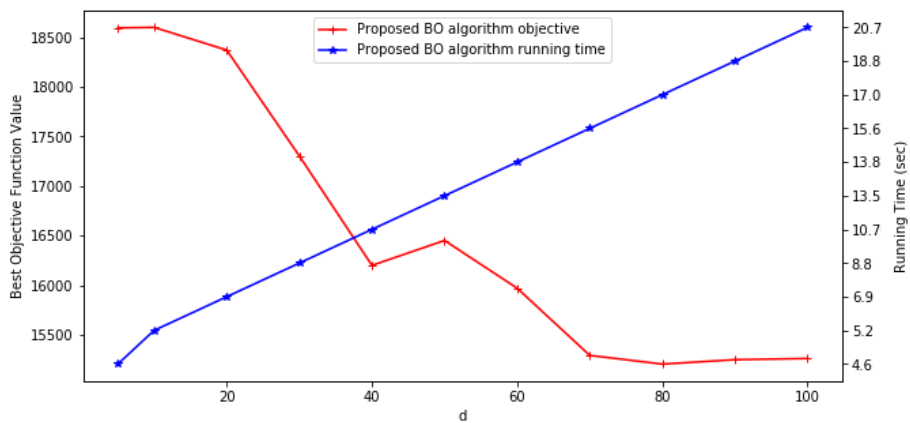

**Fig. 6.** Best objective function values and running time for different low dimension values in SEIR model

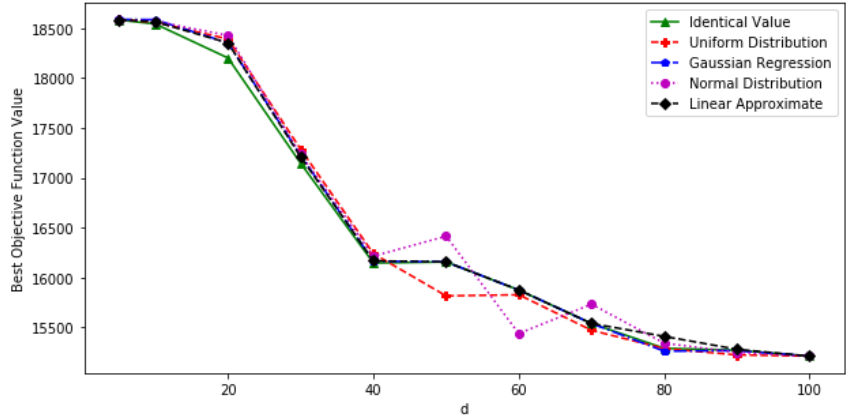

**Fig. 7.** Best objective function values for different fill-in strategies applied in deterministic SEIR model

*4.2. Global optimization on stochastic SIS epidemic optimal control model*

This part verifies the global optimization performance of the DR-DF BO algorithm on the stochastic high-dimensional SIS control model defined in Eq. (8) – Eq. (11). The stochastic SIS epidemic model has seasonal characteristics. Thus, we will study the SIS control model in a more extended period. Under this investigation, the stochastic SIS model has the control variable with 200 time-epochs, so this problem's dimensionality is 200.

**Fig. 8** shows the stochastic SIS epidemic model's comparison results with the control of different dimension reduction and without any control or interventions when the fill-in strategy is a linear approximation. The solid red line represents the trend of infectious population rate without control over time. The other color lines represent the trend of infectious population rate with optimal control generated by the DR-DF BO algorithm when the algorithm selects different low dimension values. As can be seen from **Fig. 8**, the infectious population of the stochastic SIS model has richer dynamic properties than the infectious population in the deterministic SEIR model studied in section 4.1. The trend of the epidemic performs the oscillation characteristics. The disease will come back again and again if there is not any control. However, no matter the low dimension's value, the optimal control strategies generated by the DR-DF BO algorithm effectively and quickly control the epidemic once the epidemic breakouts; it also effectively prevented the recurrence of the epidemic.

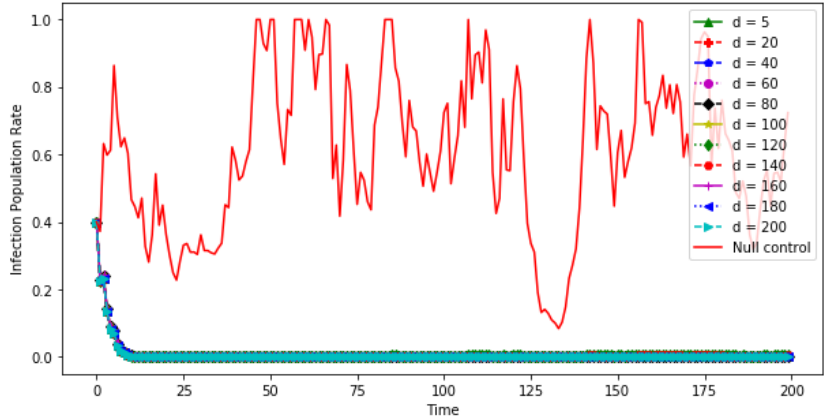

**Fig. 8.** Infectious population rate over time in stochastic SIS model for different low dimension values

**Fig. 9** shows the trends of accumulated objective function value of the stochastic SIS model with different low dimension values and without any control when the fill-in strategy is a linear approximation. The small figure in **Fig. 9** is a partial zoom figure between time interval [100,200]. We can see that the accumulated objective function when the SIS model is without control is significantly higher than the values with control. When $d$ value is about 40, the DR-DF BO algorithm can achieve a closer effect to that without dimension reduction (when $d = 200$). It means that for any low dimension values in range [40,200], the optimal control strategies generated by the DR-DF BO algorithm perform similar and sufficient control effects on the epidemic. **Fig. 10** shows the final best objective function values and running time comparison results for different low dimension values when the fill-in strategy is a linear approximation. To well understand the results, in this part we also use the AOFV ratio and RT ratio defined in Eq. (35) – Eq. (36) to present. The ratio results for the SIS model are summarized in Table 2. From the results in the table and figures, we can see that the DR-DF BO algorithm performs well both on objective function value and running time when the low dimension value is 80. It can reach good global optimization results in low dimension values is 80 with around 30 seconds running time. The DR-DF BO algorithm shows an excellent global optimization performance for the stochastic SIS model. It solves the optimal control strategy for the model in a fraction of the time. The generated control strategy can control the epidemic spread and significantly reduce the financial cost.

**Table 2. AOFV ratio and RT ratio for different $d$ in SIS model**

| $d$ | 5 | 20 | 40 | 60 | 80 | 100 | 120 | 140 | 160 | 180 | 200 |
|---|---|---|---|---|---|---|---|---|---|---|---|
| $AOFV\ ratio$ | 6.552 | 3.141 | 1.707 | 1.544 | 1.283 | 1.37 | 1.087 | 1.053 | 1.0 | 1.0 | 1.0 |
| $RT\ ratio$ | 0.146 | 0.204 | 0.288 | 0.377 | 0.461 | 0.540 | 0.648 | 0.739 | 0.813 | 0.916 | 1.0 |

**Fig. 11** shows the stochastic SIS model's best objective values with different low dimension values when the DR-DF BO algorithm chooses different fill-in strategies. As shown in **Fig. 11**, compared to the other four fill-in strategies, the Gaussian regression fill-in strategy does not perform well if the low dimension value is smaller than 40. However, these five different fill-in strategies have a good effect on global optimization if the low dimension value is larger than 80. All of them can reach an excellent and similar objective function value, which indicates that the DR-DF BO algorithm can solve the optimal control solution within a pretty short running time no matter which fill-in strategy it uses at which $d = 80$.

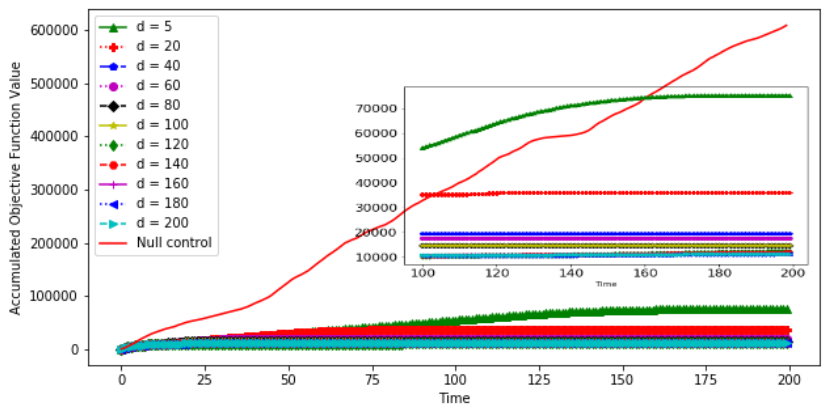

**Fig. 9.** Accumulated objective function values over time in stochastic SIS model

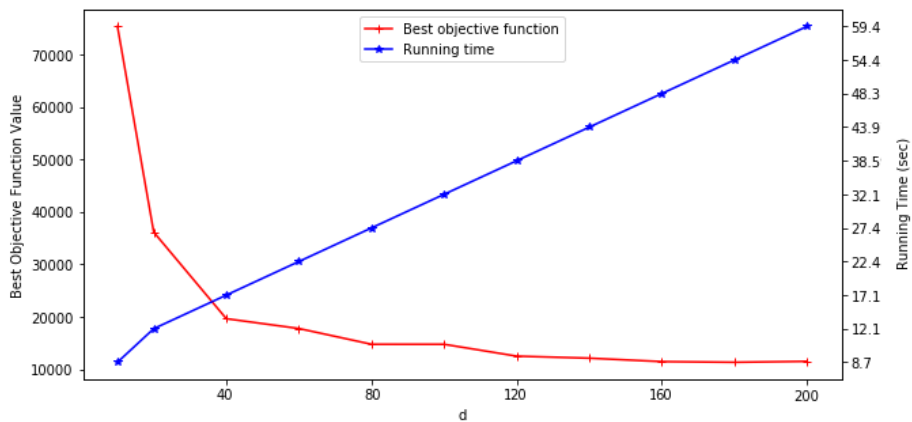

**Fig. 10.** Best objective function values and running time for different low dimension values in SIS model

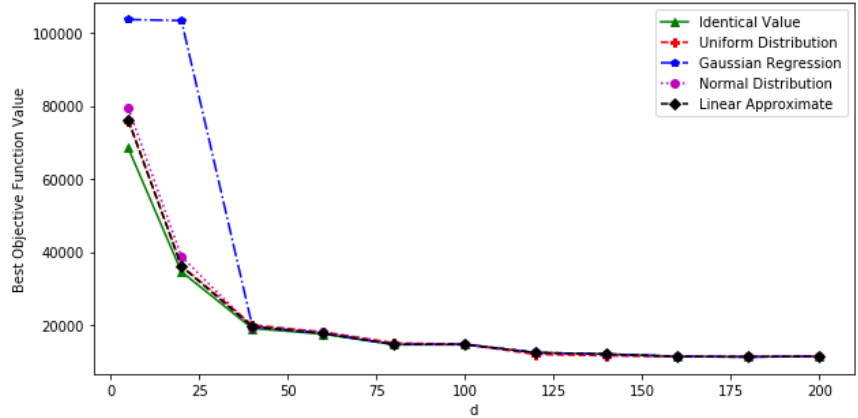

**Fig. 11.** Best objective function values for different fill-in strategies applied in stochastic SIS model

*4.3. Global optimization of the DR-DF BO algorithm compares with other Bayesian optimization algorithms*

This part compares the proposed DR-DF BO algorithm with the standard Bayesian optimization algorithm and a high dimensional Bayesian optimization algorithm proposed in [9]. We test three algorithms on the same deterministic SEIR model with the same parameter values and state conditions.

Regardless of the sampling strategy, local search, or the consideration of dimension reduction, the proposed DR-DF BO algorithm has made a significant improvement compared with the standard Bayesian optimization algorithm. Since the standard Bayesian optimization does not consider the local optimization at the final stage and dimension reduction, herein, we test the global optimization performance of the standard Bayesian optimization algorithm through different optimization iterations of the acquisition function. The result is shown in **Fig. 12**; the figure shows the final objective function values and the corresponding running time when the standard Bayesian optimization implements different iteration times to optimize the acquisition function. We can see that the final solution is sensitive to the optimization iteration times. Also, the best case of the standard Bayesian optimization algorithm can only achieve the final objective value equal to around 23,000 with about 23 seconds running time. From **Fig. 6** in **Section 4.1**, we know that the proposed DR-DF BO algorithm can reach the final objective value equal to around 18,600 with about 5 seconds running time if it considers the dimension reduction. Suppose the proposed DR-DF BO algorithm does not reduce the dimensions ($d$=100). In that case, it also achieves a better final objective function value equal to about 15,100 with 21 seconds running time, which reflects the efficiency and global optimization ability of the proposed DR-DF BO algorithm.

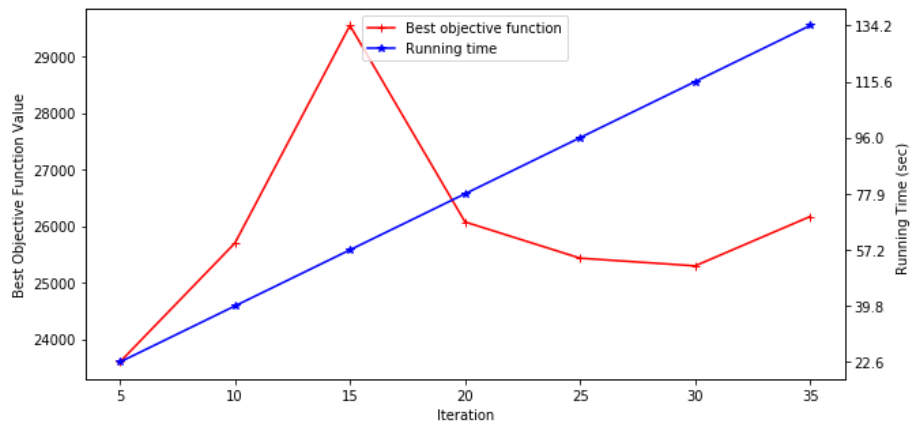

**Fig. 12.** Best objective function values and running time for different optimization iterations in SEIR model

To further demonstrate the proposed DR-DF BO algorithm's efficiency and global optimization ability, we compare it to another high dimensional Bayesian optimization algorithm in [9]. To make it easy to express, we call this comparison algorithm Referenced BO algorithm. Since the referenced algorithm considers dimension reduction, we test the two algorithms' global optimization performance for different dimension values. The results are shown in **Fig. 13**, the red lines represent the final objective function values of two algorithms for different dimension values, the blue lines represent the running time of two algorithms for different dimension values. From **Fig. 13**, we can see that regardless of the dimension value, the proposed DR-DF BO algorithm significantly performs better than the referenced algorithm in terms of the objective function value and running time.

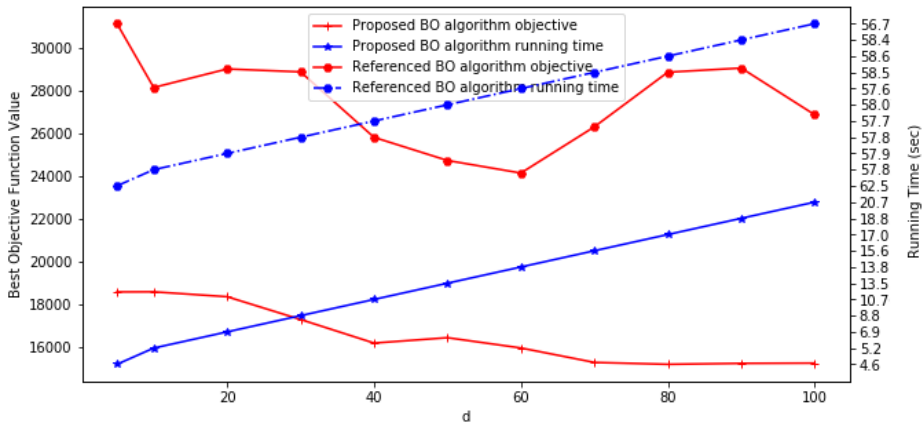

**Fig. 13.** Best objective function values and running time comparison

## 5. Conclusions

In this paper, we have proposed a high-dimensional Bayesian optimization algorithm. This algorithm is improved based on the standard Bayesian optimization algorithm. However, the proposed algorithm is effective in solving the high dimensional model. The proposed algorithm successfully implements dimension reduction and dimension fill-in with different fill-in strategies. Also, the proposed algorithm discusses a new sampling strategy to optimize the acquisition function effectively. To increase the final solution's accuracy in the high dimensional models, the proposed DR-DF BO algorithm added a local search based on a series of Adam-based steps. Comparing to the existing high dimensional Bayesian optimization algorithms, the proposed algorithm is more effective and accurate when the model is in time series.

Moreover, the proposed algorithm is not necessary to do the dimension fill-in at each acquisition function optimization process, saving time. While ensuring running time efficiency, the proposed algorithm performs better global optimization results than other high dimensional Bayesian optimization algorithms on the final optimal solution and running time. In the existing works about Bayesian optimization, researchers usually need to implement the whole process of the Bayesian optimization algorithm when it is used to solve a problem. It will waste a lot of time if the model is larger than millions of dimensions. Thus, one possible research direction would be the extension of our proposed DR-DF BO algorithm based on machine learning. It is expected to generate an optimization algorithm model containing Bayesian optimization and machine learning. After some learning process, it can solve the optimal solution for the model by only providing the initial model conditions in the future.